%% file: iclr2025_conference.tex
\documentclass{article} 
\usepackage{iclr2025_conference,times}

\input{math_commands.tex}

\usepackage{svg}
\usepackage{booktabs} 
\usepackage{hyperref}
\usepackage{bbding}
\usepackage{url}
\usepackage{multirow} 
\usepackage{xcolor}         
\usepackage{pifont}
\usepackage{graphicx}

\title{LIFT: Improving Long Context Understanding Through Long Input Fine-Tuning}

\author{Yansheng Mao$^{1,}$\thanks{Equal contributions.}, Jiaqi Li$^{2,*}$, Fanxu Meng$^{1,2}$, Jing Xiong$^{1}$, Zilong Zheng$^{2}$, Muhan Zhang$^{1,2,}$\thanks{Correspondence to Muhan Zhang (\texttt{muhan@pku.edu.cn})}\\
$^{1}$ Institute for Artificial Intelligence, Peking University \\
$^{2}$ National Key Laboratory of General Artificial Intelligence, BIGAI \\
}

\iclrfinalcopy 
\begin{document}

\maketitle

\begin{abstract}
Long context understanding remains challenging for large language models due to their limited context windows. This paper introduces \textbf{L}ong \textbf{I}nput \textbf{F}ine-\textbf{T}uning (\textbf{LIFT}) for long context modeling, a novel framework that enhances LLM performance on long-context tasks by adapting model parameters to the context at test time. LIFT enables efficient processing of lengthy inputs without the computational burden of offline long-context adaptation, and can improve the long-context capabilities of arbitrary short-context models. The framework is further enhanced by integrating in-context learning and pre-LIFT supervised fine-tuning.
The combination of in-context learning and LIFT enables short-context models like Llama 3 to handle arbitrarily long contexts and consistently improves their performance on popular long-context benchmarks like LooGLE and LongBench. We also provide a comprehensive analysis of the strengths and limitations of LIFT on long context understanding, offering valuable directions for future research.
\end{abstract}

\section{Introduction}


Large Language Models (LLMs), such as GPT-4~\citep{achiam2023gpt}, have revolutionized the field of natural language processing, driving breakthroughs in text generation and significant advancements in tasks like translation, summarization, and conversation. Lengthy sequences, which can span up to millions of tokens, are common in real-world applications including long books~\citep{kovcisky2018narrativeqa}, high-resolution videos~\citep{wu2024longmemeval, tapaswi2016movieqa}, and audio signals~\citep{yang2024air}. Extending the context window allows models to capture dependencies across larger text spans and improve coherence, understanding, and accuracy in tasks that require reasoning over extended inputs.


However, as the context length increases, the computational complexity of the self-attention mechanism~\citep{vaswani2017attention} grows quadratically, which limits the model's ability to process long inputs. Additionally, storing a large number of attention weights and intermediate states places a heavy burden on hardware resources. Moreover, it's challenging to capture long dependencies among pieces of information scattered throughout raw texts and perform further comprehension and reasoning. Due to the limitation of context window, LLMs can hardly capture the overall information about a user's query history or task input, resulting in suboptimal performance.

To address these challenges, researchers have developed various techniques to improve the long-context abilities of LLMs.
A line of research, including Retrieval-Augmented Generation (RAG)~\citep{lewis2020retrieval, xu2023retrieval} and prompt compression~\citep{jiang2023longllmlingua}, preprocesses inputs and provide a short sequence to LLMs~\citep{el2021automatic}. However, the effectiveness of these methods depends on the precision and relevance of the contextual information provided within the context window. 
When limited, ambiguous, or conflicting information is provided in the context window, it can lead to hallucination.
Another line of research, long-context adaptation, focuses on fine-tuning pretrained LLMs on corpora of long texts to extend their context windows~\citep{chen2023longlora, peng2023yarn}. However, it comes with significant costs in terms of training data and computational resources.
Additionally, with the extended context window, the cost of processing and generating long texts also grows quadratically. Finally, despite the extension, the context window of these LLMs remain limited, preventing them from generalizing to inputs of infinite length.


Therefore, in this paper, we present a novel framework \textbf{L}ong \textbf{I}nput \textbf{F}ine-\textbf{T}uning (LIFT), designed to enhance the long-context capabilities of any short-context model by directly adapting model parameters to the long input. Our approach has the following advantages:
\begin{itemize}
    \item \textbf{Efficient long-input training on the fly.} LIFT dynamically adapts to long inputs by tuning model parameters, eliminating the need for resource-intensive offline long-context adaptation and expensive long-context inference. Incorporating segmentation strategies tackles the problems of long text fine-tuning with short context window.
    \item \textbf{Enhanced long-context in-context learning (ICL) capability with LIFT.} LIFT provides an efficient approach to long-context comprehension by complementing ICL. For tasks requiring the integration of new knowledge, the introduction of truncated ICL further enhances LIFT's ability to dynamically utilize relevant information, improving adaptability and performance.
    \item \textbf{Great improvement on certain long context tasks.} Based on our evaluations on various benchmarks, LIFT shows substantial benefits for basic tasks such as summarization, as well as more complex tasks such as timeline reordering and reading comprehension. It has been validated that integrating pre-LIFT supervised fine-tuning on similar long inputs helps to further strengthen the model's capability on downstream tasks.
\end{itemize}

We conduct a comprehensive evaluation of LIFT under various settings. The results demonstrate that, when integrated with ICL, LIFT significantly enhances performance on tasks such as timeline reordering in the LooGLE~\citep{li2023loogle} benchmark, as well as NarrativeQA, QMSum, and GovReport in LongBench~\citep{bai2023longbench}, showcasing its potential for generalization across diverse tasks. These findings highlight the effectiveness of LIFT in improving the contextual comprehension of short-context models, paving the way for broader applications in long-context scenarios.

\section{Related works}

\paragraph{Long context adaptation and efficient architectures.} One conventional approach for handling long contexts is to place all inputs into the context and leverage in-context learning (ICL). However, short-context models fail to generalize to long contexts due to unseen positional encodings, resulting in poor performance on long-context tasks. To address this, many studies fine-tune LLMs on corpora of long texts to extend their context window. While this approach effectively enhances long-context understanding, it comes at the expense of efficiency, as both long-context adaptation and long-context ICL are computationally expensive.

To speed up long-context processing, one popular approach is developing efficient Transformers. Sparse attention~\citep{kitaev2020reformer, wang2020linformer, beltagy2020longformer} reduces memory and computation costs by using techniques like local windows or strided attention, allowing models to focus on relevant parts of the input. Linear attention~\citep{shen2021efficient} reduces complexity from quadratic to linear by approximating self-attention with kernel functions or low-rank representations.

Another direction involves Transformer alternatives, such as state-space models (SSMs), which are efficient in both training and inference due to their dual representations. However, these methods often reduce expressiveness and are less adopted by modern LLMs. In this work, we focus on the conventional self-attention mechanism~\citep{vaswani2017attention}.

\paragraph{Retrieval-Augmented Generation (RAG).} RAG~\citep{lewis2020retrieval}  improves the efficiency and effectiveness of LLMs in long-context understanding by integrating external memory components~\citep{xu2023retrieval, jiang2024longrag, wang2024augmenting, jin2024llm}. It stores information over time, allowing the model to recall past information without requiring the entire context to fit within its context window. However, the retrieval quality is a bottleneck of RAG. Inaccurate or noisy retrieval leads to performance degradation and hallucination.

\paragraph{Test-time training for long-context understanding} Test-time training (TTT)~\citep{liu2021ttt++, gandelsman2022test, osowiechi2023tttflow} has emerged as a promising approach to adapt models to unseen data distributions during deployment, leveraging the test data to fine-tune the model at inference time. \cite{sun2020test} introduced TTT by utilizing self-supervised tasks on test inputs to refine the model in real-time, ensuring better generalization to out-of-distribution data. Subsequent research has expanded this idea to enhance robustness against distribution shifts, adversarial attacks, and noisy environments. 

Recent works have applied TTT to improve model adaptability when dealing with lengthy, context-rich inputs. \cite{sun2024learning} proposed new TTT layers by updating the hidden states through self-supervised learning, significantly improving performance over traditional RNNs and matching Transformer models in processing long texts up to 16k tokens. \cite{bertsch2024context} studied the benefits of TTT in improving in-context learning by dynamically refining representations based on extended input sequences. \cite{wang2024greater} explores how TTT can enhance LLMs in long-text generation tasks such as novel writing and translation. 
Our work focuses on improving arbitrary models' long-context capabilities by fine-tuning the model on the long input, which is not restricted to specific models or layers.

\section{Method}
\label{sec:method}

\begin{table}[t]
    \centering
    \caption{Comparison between conventional approaches and LIFT.}
    \label{tab:compare}
    {\begin{tabular}{l|ccc}
    \toprule
     & RAG & ICL & LIFT \\
    \midrule
    Knowledge storage & External data sources & Within context window & In parameters \\
    Input size &Infinite & Limited & Infinite \\
    Retrieval & \textcolor{green}{\ding{51}} & \textcolor{red}{\ding{55}}  &\textcolor{red}{\ding{55}} \\
    Long-context adaptation & \textcolor{red}{\ding{55}} & \textcolor{green}{\ding{51}}  &\textcolor{red}{\ding{55}} \\
    Long-context inference &\textcolor{red}{\ding{55}} & \textcolor{green}{\ding{51}}  &\textcolor{red}{\ding{55}} \\
    \bottomrule
    \end{tabular}}
    \vspace{-8pt}
\end{table}



As discussed earlier, our method features adaptation and inference with only a short-context model, ensuring high efficiency. The comparison of our method, LIFT, with other long-context processing methods, ICL and RAG, is illustrated in Table~\ref{tab:compare}. In this section, we present how we implement LIFT and address the associated challenges.

\begin{itemize}
    \item In Section \ref{subsec:segment}, we introduce our basic setup, which involves training on segments of long input.
    \item In Section \ref{subsec:AT}, we compensate for potential capability loss and enable the model to perform reasoning over long input by incorporating auxiliary tasks (AT) during fine-tuning.
    \item In Section \ref{subsec:SFT}, we further refine the model by supervised fine-tuning it on a diverse set of long documents and synthetic tasks, making it familiar with our LIFT paradigm and adapts to new long texts better.
\end{itemize}

\subsection{Training with input segments}
\label{subsec:segment}

LLMs access knowledge either from contexts or their parameters. Unlike ICL, we propose storing test-time knowledge in the parameters by adapting the model to the given long input.

We formalize memorizing the input as a language modeling task. Let the input be $\mathbf{x}=(x_{1},x_{2},\dots,x_{L})$, where $L$ is a very large number. The objective function for the language modeling task is defined as
$$
\mathcal{L}_{LM}(\mathbf{x};\theta)=\sum_{i=1}^{L}\log\mathbb{P}(x_{i}|\mathbf{x}_{1:i-1};\theta),
$$
where $\theta$ is the parameters.

\begin{figure}
    \centering
    \includegraphics[width=0.75\linewidth]{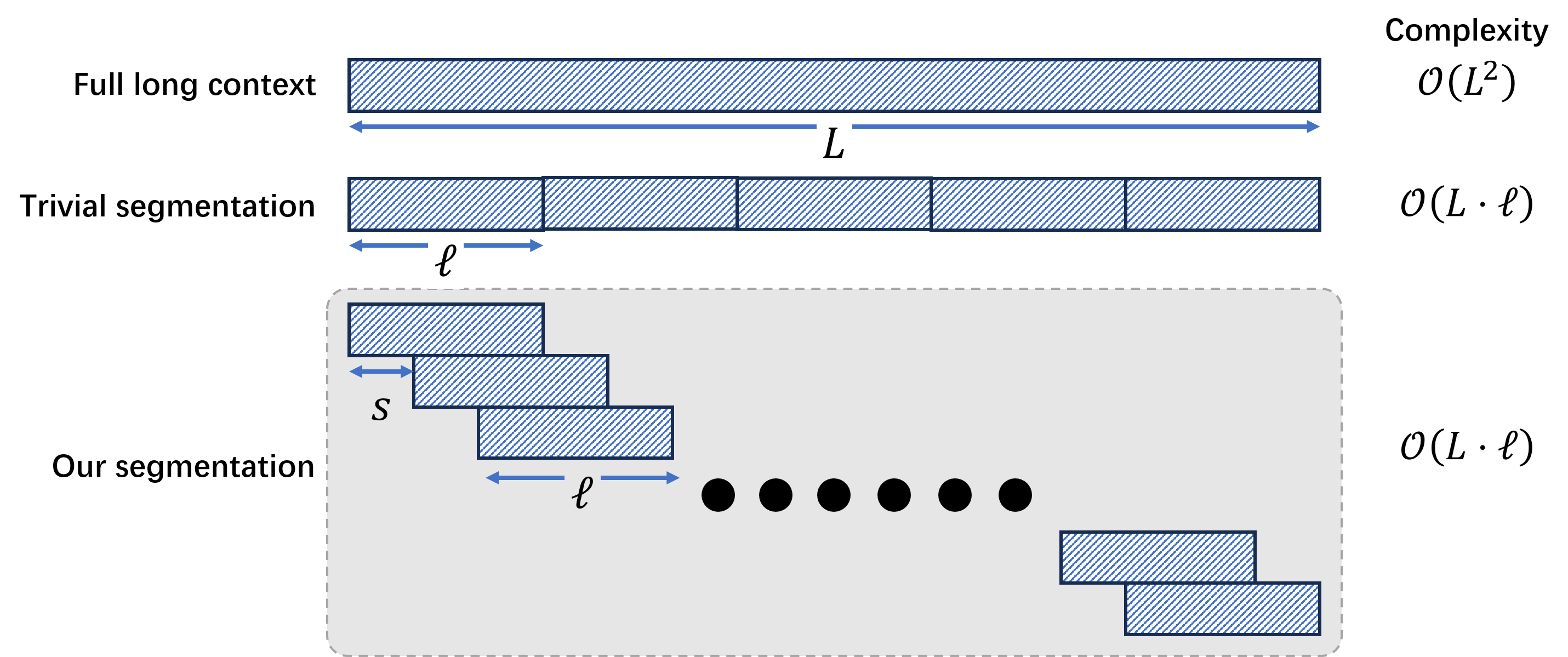}
    \caption{Comparison between our segmentation method and the trivial segmentation method.}
    \label{fig:segmentation}
\end{figure}

However, directly adapting the model to a long text of length $L$ incurs a computational complexity of $\mathcal{O}(L^{2})$ and becomes infeasible when the base model has a context window shorter than $L$. A straightforward approach is to truncate $\mathbf{x}$ into non-overlapping short segments, denoted as $\mathbf{x}_{l_{1}:r_{1}},\dots,\mathbf{x}_{l_{K}:r_{K}}$, as illustrated in Figure \ref{fig:segmentation} (Trivial segmentation). The objective function for the language modeling task with the short segments is expressed as
$$
\mathcal{L}_{input}(\mathbf{x};\theta)=\sum_{k=1}^{K}\mathcal{L}_{LM}(\mathbf{x}_{l_{k}:r_{k}};\theta).
$$

However, the trivial segmentation fails to preserve the sequential order of the segments. Since there is no overlap between the adjacent segments, the model cannot infer the correct order of the segments.

To address this issue, we propose an intuitive solution: introducing overlaps between the adjacent segments, as illustrated in Figure \ref{fig:segmentation} (Our segmentation). By overlapping the tail of one segment with the head of the next, the model can better retain the sequential order of the context. Ideally, if the model learns to generate the tail of one segment, it can continue to recite the next segment. Formally, we design that
$$
\begin{aligned}
&l_{1}=1,r_{K}=L,\\
&\forall i=1,2,\dots,K-1,r_{i}-l_{i}+1=\ell,l_{i+1}=l_{i}+s.
\end{aligned}
$$

Here $s$ controls the length of the overlaps. Empirically, taking $s=\frac{3}{8}\ell$ proves sufficient in our experiments, which introduces only constant computational complexity overhead.

\subsection{Training with auxiliary tasks}
\label{subsec:AT}





Adapting a pretrained LLM to a specific task risks damaging its other capabilities. Similarly, while adapting to the input helps the model memorize the input, it probably degrades other abilities, such as instruction-following. Moreover, effectively memorizing the long input doesn't mean the model can reason based on it.

To mitigate potential capability loss and enable the model to reason based on the long context, we propose synthesizing auxiliary question-answering (QA) tasks, denoted as $(\mathbf{q}_{i},\mathbf{a}_{i})_{i=1}^{m}$, based on the long context. The objective function of the auxiliary tasks is defined as
$$
\mathcal{L}_{AT}((\mathbf{q}_{i},\mathbf{a}_{i})_{i=1}^{m};\theta)=-\sum_{i=1}^{m}\log\mathbb{P}[\mathbf{a}_{i}\mid\mathbf{q}_{i};\theta].
$$

Following the mechanism of mix training~\citep{AllenZhu2023PhysicsOL}, which asserts that LLMs can only learn to perform inference based on $\mathbf{x}$ when trained simultaneously on both $\mathbf{x}$ and $(\mathbf{q}_{i},\mathbf{a}_{i})_{i=1}^{m}$, we propose jointly optimizing the two objective functions, i.e.,
$$
\mathcal{L}(\mathbf{x},(\mathbf{q}_{i},\mathbf{a}_{i})_{i=1}^{m};\theta)=\mathcal{L}_{input}(\mathbf{x};\theta)+\gamma\cdot\mathcal{L}_{AT}((\mathbf{q}_{i},\mathbf{a}_{i})_{i=1}^{m};\theta).
$$

There are no strict constraints on the method used to synthesize $(\mathbf{q}_{i},\mathbf{a}_{i})_{i=1}^{m}$ based on $\mathbf{x}$, except that is should avoid computationally expensive operations on $\mathbf{x}$, such as inference over the entire $\mathbf{x}$. In our experiments, we extract several short segments from $\mathbf{x}$ and use a pretrained LLM to generate QA pairs based on the segments.

\subsection{Further improvement with pre-LIFT Supervised Fine-Tuning}
\label{subsec:SFT}

While our framework LIFT is applicable to any model capable of fine-tuning, we suggest that pretrained LLMs may be unfamiliar with our training method, which leads to suboptimal results. We hypothesize that performance on downstream tasks can be enhanced by learning a new set of parameters through multiple rounds of LIFT with auxiliary tasks, a process commonly known as Supervised Fine-Tuning (SFT), which has been shown to be effective for long-context downstream tasks~\citep{beltagy2020longformer,zaheer2021bigbirdtransformerslonger}. Based on this SFT model, we will then apply the normal LIFT process to further adapt the model to the given test input.

The SFT process involves training the model on a large corpus of long texts, combined with QA tasks synthesized based on the corpus. To ensure the model becomes familiar with our LIFT method, the supervised fine-tuning (SFT) tasks are designed to closely resemble those used in our LIFT framework. Unlike our main approach, where the model adapts to a single piece of long text, the SFT phase involves adapting the model to multiple pieces of long text simultaneously, preventing it from overfitting.

Formally, we select the corpus $(\mathbf{x}^{(i)})_{i=1}^{N}$ independent of the test datasets. For each $\mathbf{x}^{(i)}$, we synthesize a set of QA tasks $(\mathbf{q}_{j}^{(i)},\mathbf{a}_{j}^{(i)})_{j=1}^{K}$. The objective function for SFT is defined as
$$
\mathcal{L}_{SFT}\Big(\big(\mathbf{x}^{(i)},(\mathbf{q}_{j}^{(i)},\mathbf{a}_{j}^{(i)})_{j=1}^{K}\big)_{i=1}^{N};\theta\Big)=\frac{1}{N}\sum_{i=1}^{N}\left(\mathcal{L}_{input}(\mathbf{x}^{(i)};\theta)+\gamma\cdot\mathcal{L}_{AT}((\mathbf{q}_{j}^{(i)},\mathbf{a}_{j}^{(i)})_{j=1}^{K};\theta)\right).
$$

\begin{figure}[t]
    \centering
    \includegraphics[width=0.9\linewidth]{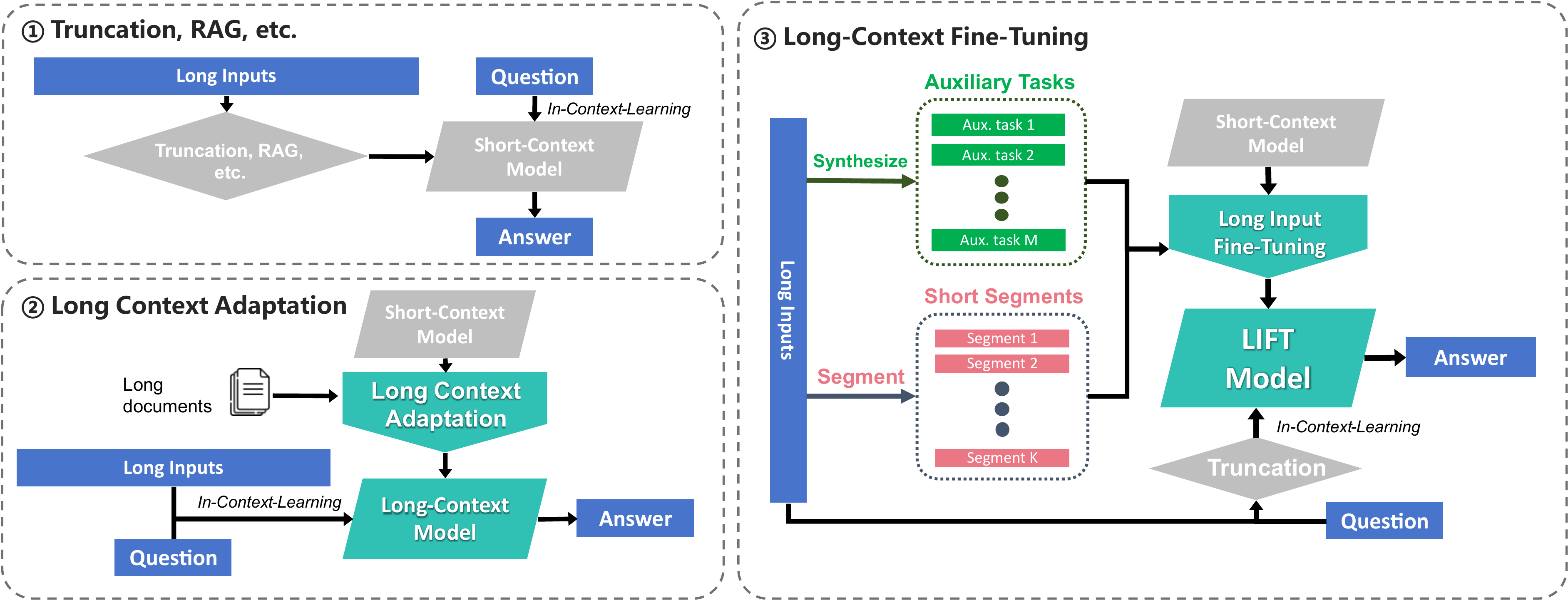}
    \caption{An overview of our method compared with existing methods like truncation, RAG, and long context adaptation.}
    \label{fig:method}
\end{figure}

An overview of our LIFT (in comparison to other mainstream long-context approaches) is presented in Figure~\ref{fig:method}.

\section{Experiments}
\label{others}

\subsection{Setup}

\paragraph{Dataset and metrics}
To evaluate our method, we choose three popular long-context benchmarks, including LooGLE~\citep{li2023loogle}, LongBench~\citep{bai2023longbench}, BAMBOO~\citep{dong2023bamboo} and Quality.
They provide a relatively comprehensive evaluation, covering a wide variety of application scenarios. The evaluation metrics are task-specific and consistent with the respective original benchmarks~\citep{banerjee-lavie-2005-meteor, zhang2020bertscoreevaluatingtextgeneration}. Among these, the GPT-4 score evaluates the correctness of responses of LLMs given corresponding questions and answers with GPT-4, which is proven to be highly aligned with human evaluations.

\paragraph{Models}
For open-source LLMs, we select LLaMA3-8B-Instruct~\citep{llama3modelcard} with 8k context window. For closed-source commercial LLMs, we choose GPT3.5-turbo-16k~\citep{chen2023robust} with 16k context window. It has shown competitive performance on popular long context benchmarks and can be accessed for further fine-tuning. Details of the models and training parameters used can be seen in Appendix \ref{app:details}.

\paragraph{Settings}
In our main results, we compare four methods below as different settings for comparison.
\begin{itemize}
    \item ICL with truncation (noted as \texttt{ICL}), where we truncate the input by only keeping its beginning and end tokens to maximally fill the context window, and use the original LLM.
    \item LIFT without ICL (noted as \texttt{LIFT\_only}), where we use the LIFT LLM without filling any input into the context window.
    \item LIFT with ICL (noted as \texttt{LIFT+ICL}), where we use the LIFT LLM and additionally fill the beginning and end tokens of the input tokens into the context window.
\end{itemize}

By default, LIFT does not use auxiliary tasks (AT) and SFT and only adapts the model to the input text.

\subsection{Main results}
\label{sec:main_results}
\subsubsection{Results on LooGLE}
\label{sec:main_results_loogle}
\paragraph{Overall performance.} As shown in Table \ref{tab:main_result}, \texttt{LIFT+ICL} consistently achieves the highest scores across both LongQA and ShortQA tasks for both models, and is particularly effective in the ShortQA task, which doesn’t rely on long dependencies. Interestingly, \texttt{LIFT\_only} performs the worst among all the settings.


Compared to GPT-3.5, Llama 3 benefits more from \texttt{LIFT+ICL}, showing notable improvement in GPT4\_score: from 30.88 (\texttt{ICL}) to 33.42 in LongQA, and from 44.23 to 50.44 in ShortQA. These results highlight that LIFT significantly improves the performance of ICL, particularly for models with short context windows. Notably, GPT-3.5 generally outperforms Llama 3 across the tasks, especially in ShortQA, where it achieves a GPT4\_score of 69.66 compared to 50.44 of Llama 3.
Notably, all models perform particularly poorly on LongQA, with GPT4\_score falling below 50. This underscores that modeling long dependencies in extended contexts remains a significant challenge for existing models.

\begin{table}[t]
    \centering
    \caption{Performance on LooGLE under different settings}
    \label{tab:main_result}
    {\begin{tabular}{ccc|ccc}
    \toprule
    \multicolumn{1}{c}{Models} & 
    \multicolumn{1}{c}{Task} & 
    \multicolumn{1}{c|}{Methods} & 
    \multicolumn{1}{c}{Meteor} & 
    \multicolumn{1}{c}{Bertscore} & 
    \multicolumn{1}{c}{GPT4\_score} \\
    \midrule
     \multicolumn{1}{c}{\multirow{6}{*}{\textbf{LLaMa3}}} 
     & \multicolumn{1}{c}{\multirow{2}{*}{\textbf{LongQA}}} & \texttt{ICL} & 9.10 & 83.60 & 30.88 \\
     & & \texttt{LIFT\_only} &9.07 & 83.47 & 27.34\\
     & & \texttt{LIFT+ICL} & \textbf{9.15} &\textbf{83.71} &\textbf{33.42} \\
    \cmidrule(lr){2-6}
    &  \multicolumn{1}{c}{\multirow{2}{*}{\textbf{ShortQA}}} & \texttt{ICL} & 23.45  & 86.01 &44.23 \\
     & & \texttt{LIFT\_only} &18.29  &85.21 &35.83 \\
     & & \texttt{LIFT+ICL} &\textbf{24.06} & \textbf{86.13} &\textbf{50.44} \\
     \midrule
     \multicolumn{1}{c}{\multirow{6}{*}{\textbf{GPT3.5}}} 
     & \multicolumn{1}{c}{\multirow{2}{*}{\textbf{LongQA}}} & \texttt{ICL} & 11.71 & 85.48 &44.82 \\
     & & \texttt{LIFT\_only} & 9.95 & 85.33 & 35.22 \\
     & & \texttt{LIFT+ICL} & \textbf{11.74} & 85.48 & \textbf{45.76} \\
    \cmidrule(lr){2-6}
    &  \multicolumn{1}{c}{\multirow{2}{*}{\textbf{ShortQA}}} & \texttt{ICL} & 32.58 & 87.04 & 66.82 \\
     & & \texttt{LIFT\_only} & 18.53  & 86.63 & 33.98 \\
     & & \texttt{LIFT+ICL} & \textbf{37.38}  & \textbf{88.73} & \textbf{69.66}\\
    \bottomrule
    \end{tabular}}
    \vspace{-8pt}
\end{table}

\begin{table}[t]
    \centering
    \caption{Performance of each LongQA task in LooGLE using GPT4\_score}
    \label{tab:main_result_task}
    \resizebox{\linewidth}{!}{\begin{tabular}{ll|cccc}
    \toprule
    Models & Methods & Comprehension  \& Reasoning & Multiple info retrieval & Computation & Timeline reorder \\
     \midrule
     \multicolumn{1}{c}{\multirow{2}{*}{\textbf{LLaMa}}} 
     & \texttt{ICL}  & 40.88 & 28.16 & \textbf{24} & 22.33\\
     & \texttt{LIFT+ICL} & \textbf{44.83} & 28.16 & 22 & \textbf{26.51}  \\
    \midrule
    \multicolumn{1}{c}{\multirow{2}{*}{\textbf{GPT3.5}}} 
     & \texttt{ICL} & 52.67 & \textbf{40.77}  & \textbf{27.55} & 45.19\\
     & \texttt{LIFT+ICL} & \textbf{53.44} &40.50 & 26.53 &\textbf{49.52}\\
    \bottomrule
    \end{tabular}}
    \vspace{-8pt}
\end{table}

\paragraph{Performance on each LongQA task.} Table \ref{tab:main_result_task} presents further experimental results across four LongQA tasks introduced in LooGLE. While GPT-3.5 consistently shows significant advantages over Llama 3 as shown in Table \ref{tab:main_result}, Llama 3 exhibits greater relative improvement from LIFT in certain tasks, particularly in Comprehension \& Reasoning and Timeline Reordering.

In Comprehension \& Reasoning, Llama 3 achieves a score of 40.88 with \texttt{ICL} and shows a notable improvement to 44.83 with \texttt{LIFT+ICL}. Similarly, on Timeline Reordering, it demonstrates a significant improvement, reaching 26.51 with \texttt{LIFT+ICL}. These results reveal that LIFT enhances ICL for both models by enabling a more comprehensive understanding of the entire lengthy input, which is effectively encoded in the parameters.

However, LIFT does not show any improvement in tasks such as Multiple Information Retrieval and even results in a slight performance degradation in Computation for both models. This indicates that LIFT may not uniformly benefit all tasks and, in some cases, could introduce noise. Its effectiveness appears to vary depending on the task and the inherent strengths of the model, highlighting the need for task-specific considerations when applying LIFT.

\subsubsection{Results on LongBench}
\label{sec:main_results_longbench}

Table \ref{tab:main_result_longbench} presents the results across five tasks in LongBench.
Those tasks are selected as representatives of the English-based tasks with sufficiently long inputs.
We keep the metrics the same as the original benchmark.

Overall, GPT-3.5 with longer context window outperforms Llama 3 on all tasks using the same method except GovReport. However, Llama 3 gains greater benefits from \texttt{LIFT+ICL} than GPT-3.5, which is consistent with the results on LooGLE, suggesting that LIFT is especially helpful for models with short context window. \texttt{LIFT+ICL} consistently outperforms both \texttt{ICL} and \texttt{LIFT\_only} on Narrativeqa and Qmsum for both models showing its advancement in long context understanding. Notable improvements are seen in Narrativeqa from 20.73 to 25.84.

For Musique and GovReport, it shows a different trend on the two models. Llama 3 shows slight improvements on GovReport while its performance drops a lot on Musique. GPT-3.5 demonstrates the opposite pattern. Notably, the performance of Llama 3 on PassageRetrievalEN significantly drops when using \texttt{LIFT+ICL} compared to \texttt{ICL}, indicating that LIFT's effectiveness varies across tasks. This encourages us to leverage LIFT's potential through task-level tuning.

\begin{table}[t]
    \centering
    \caption{Performance on LongBench under different settings}
    \label{tab:main_result_longbench}
    {\begin{tabular}{cc|ccccc}
    \toprule
    \multicolumn{1}{c}{Models} & 
    \multicolumn{1}{c|}{Methods} & 
    \multicolumn{1}{c}{Musique} & 
    \multicolumn{1}{c}{Narrativeqa} & 
    \multicolumn{1}{c}{Qmsum} & 
    \multicolumn{1}{c}{GovReport} & 
    \multicolumn{1}{c}{PassageRetrievalEN} \\
    \midrule
     \multicolumn{1}{c}{\multirow{3}{*}{\textbf{LLaMa3}}} 
      & \texttt{ICL} & \textbf{15.61} & 20.73 & 21.28 &29.39 & \textbf{61.11} \\
     &  \texttt{LIFT\_only} &5.21 & 5.55 & 14.00 & 12.50 & 9.03 \\
     &  \texttt{LIFT+ICL} &10.99 & \textbf{25.84} & \textbf{22.96} &\textbf{31.26} & 41.67 \\
     \midrule
     \multicolumn{1}{c}{\multirow{3}{*}{\textbf{GPT3.5}}} 
     & \texttt{ICL} & 26.33  & 25.67 & 22.09 &\textbf{25.30} & \textbf{79.17}\\
     & \texttt{LIFT\_only} & 13.58 & 11.95 & 18.76 & 8.90 & 6.25 \\
     & \texttt{LIFT+ICL} & \textbf{27.20} & \textbf{26.53} & \textbf{22.23} & 25.01 & \textbf{79.17}\\
    \bottomrule
    \end{tabular}}
    \vspace{-8pt}
\end{table}

\subsubsection{Efficiency}

Benefiting from our truncation strategy (Section \ref{subsec:segment}), the computational complexity of our method scales linearly with the input context length. To further evaluate the efficiency of our approach compared to ICL, we measure the time cost of a single Needle-In-A-Haystack (NIAH) task under both methods. In this experiment, the input lengths are controllable and the primary computational cost stems from processing the input context rather than iterative generation.

\begin{figure}
    \centering
    \includegraphics[width=0.8\linewidth]{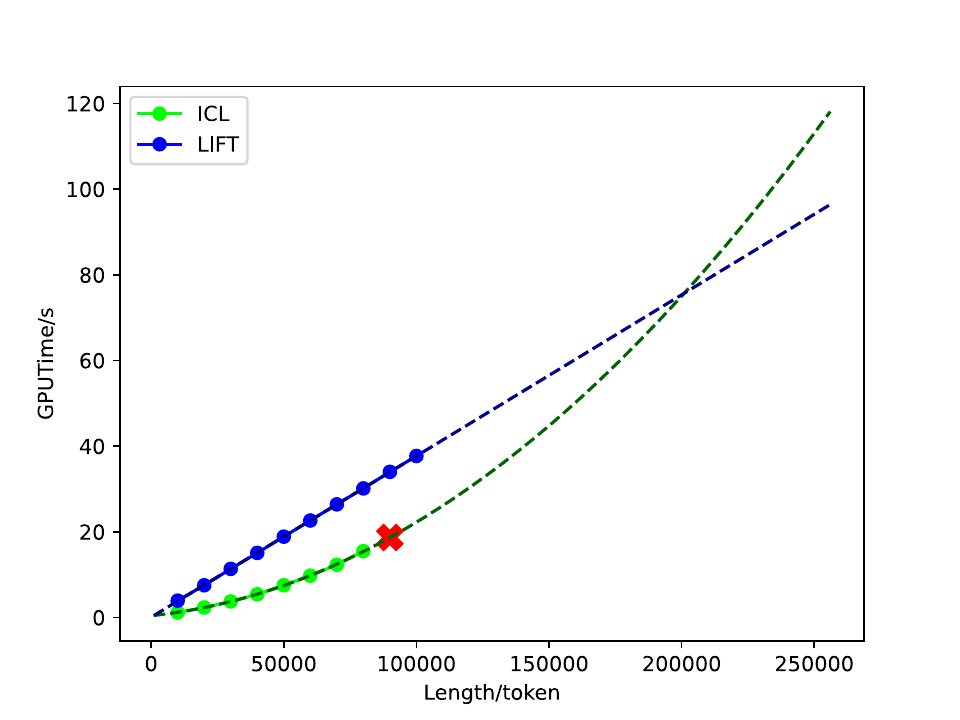}
    \caption{GPU time vs. input length for LIFT and ICL. The dashed lines represent the fitted curves, showing linear growth for LIFT and quadratic growth for ICL. The red cross indicates the input length at which ICL runs out of memory.}
    \label{fig:efficiency}
\end{figure}

We plot the GPU time against the input length along with the fitted curves in Figure \ref{fig:efficiency}.

First, we observe that LIFT is significantly more memory-efficient than ICL. Notably, ICL runs out of memory when the input length exceeds 90k tokens on our A100 (80G) system. Upon closer inspection, we find that the cache of hidden states for previous tokens consumes most of the memory in ICL. In contrast, LIFT is capable of handling arbitrarily long inputs. Our truncation strategy ensures that LIFT only involves adaptation and inference with short text segments, eliminating the need for extensive caching.

Empirically, we find that the time cost of ICL grows quadratically with input length, while our method scales linearly. However, we also observe that the constant factor introduced by adaptation in the computational complexity of LIFT is non-negligible. As a result, our method only surpasses ICL in time efficiency when the input length exceeds a certain threshold above 200k tokens. The primary cost of our method arises from the multi-epoch fine-tuning. We hypothesize that by using better parallel fine-tuning techniques and designing tasks that are more aligned with the strengths of LIFT, the efficiency of the LIFT framework can be significantly improved.


\subsection{Further studies on enhancing LIFT capability}
\label{sec:further_studies}

Encouraged by the significant improvement observed in the timeline-reorder task from LooGLE, we aim to further enhance the performance of LIFT on similar tasks like sorting and reordering, by incorporating auxiliary tasks (AT, Section \ref{subsec:AT}) and pre-LIFT SFT (Section \ref{subsec:SFT}). For AT, we generate synthetic QAs according to the input text simliar to the target task and fine-tunes the model on both the input text and the QAs. For SFT, we generate synthetic QAs on independent corpus and fine-tune the model on the corpus and QAs before applying LIFT on specific inputs. 

The results are illustrated in Table \ref{tab:sft_at_result}. There are six models compared:
\begin{itemize}
    \item \texttt{ICL} and \texttt{LIFT+ICL} are the same as before;
    \item \texttt{LIFT+AT+ICL} means fine-tuning on both input text and synthetic QAs during the LIFT phase;
    \item \texttt{SFT+ICL}, \texttt{SFT+LIFT+ICL} and \texttt{SFT+LIFT+AT+ICL} mean using the SFT model rather than the original LLM for the previous three baselines.
\end{itemize}

Comparing the results of \texttt{LIFT+ICL} and \texttt{LIFT+AT+ICL}, as well as \texttt{SFT+LIFT+ICL} and \texttt{SFT+LIFT+AT+ICL}, we observe that AT brings negligible improvement or even slightly degrades performance for the LIFT phase. A possible explanation is that the number of synthesized samples in our evaluation is insufficient, potentially causing the model to overfit these specific examples instead of enhancing the general ability. However, it's impractical to synthesize a huge number of training samples at test time due to unacceptable computational cost. Striking a balance between efficiency and effectiveness when using AT at test time remains a significant challenge and requires further exploration.

In contrast, we find SFT greatly improves the performance of both ICL and LIFT+ICL, which is reasonable since the tasks used in the SFT process are similar to those at test time. \texttt{SFT+LIFT+ICL} is still better than \texttt{SFT+ICL}, highlighting the effectiveness of LIFT.


\begin{table}[t]
    \centering
    \caption{Coordinate score on specific task in Bamboo, LooGLE, and QuALITY using AT and SFT.}
    \label{tab:sft_at_result}
    \resizebox{0.6\linewidth}{!}{
        \begin{tabular}{l|ccc}
        \toprule
        Methods                     & Bamboo            & LooGLE            & QuALITY           \\
        \midrule
        \texttt{ICL}                & 29.03             & 19.97             & 24.06             \\
        \texttt{LIFT+ICL}           & 30.83             & 21.04             & 24.19             \\
        \texttt{LIFT+AT+ICL}        & 30.90             & 21.14             & 23.75             \\
        \texttt{SFT+ICL}           & 31.00             & \textbf{24.11}             & 25.23             \\
        \texttt{SFT+LIFT+ICL}       & \textbf{32.47}    & 23.52             & \textbf{25.48}    \\
        \texttt{SFT+LIFT+AT+ICL}    & 31.13             & 22.13             & 24.57             \\
        \bottomrule
        \end{tabular}
    }
    \vspace{-8pt}
\end{table}








\section{Conclusion}

In this paper, we proposed a novel framework, \textbf{L}ong-\textbf{I}nput \textbf{F}ine-\textbf{T}uning (\textbf{LIFT}), to enhance LLMs' long-context understanding. Our approach dynamically adapts to long inputs by efficiently fine-tuning the model parameters and utilizing the in-parameter knowledge to improve long-context understanding. Experimental results across popular benchmarks like LooGLE and LongBench demonstrate that the combination of ICL and LIFT enables short-context models to solve long-context tasks with great improvement on some long-context tasks. In particular, LIFT is significantly more memory efficient than conventional ICL.

\section{Limitations and future works}
\label{sec:limitation}

\paragraph{Limitations of LIFT without ICL.} While we often employ truncated contexts to simplify inference on lengthy texts, this approach is proven insufficient for tasks that demand precise information extraction from extended contexts, such as the Needle in a Haystack (NIAH) task. Despite the practical value of NIAH is arguable, we still perform the experiments and show the results in Appendix \ref{app:niah}. For NIAH tasks, \texttt{LIFT\_only} is insufficient and ICL using a long context seems indispensable.

\paragraph{More advanced LIFT methods.} We introduce an intuitive strategy, LIFT, for handling long contexts, showcasing its potential to address challenges associated with lengthy inputs. However, pretrained LLMs may not be naturally familiar with the LIFT framework. To bridge this gap, we introduce pre-LIFT SFT, but our vision is to generalize the LIFT framework to any pretrained LLM, enhancing its flexibility and adaptability without requiring extensive retraining. This still needs extensive future study.

\paragraph{Strategy to extract parametric knowledge after LIFT} Through LIFT, embedding the inputs into the model's internal parameters enhances its familiarity with the inputs. However, the effectiveness of downstream tasks still depends on the model's ability to autonomously extract and utilize the parametric knowledge gained during LIFT. Our experiments (Appendix B) reveal that explicitly providing task-relevant knowledge outperforms using LIFT alone. Furthermore, supplying task-relevant knowledge to the model after applying LIFT still significantly improves the performance. This underscores the potential of developing strategies to effectively trigger and leverage LIFT-acquired knowledge for downstream tasks (such as using RAG), making it a promising direction for further research and exploration.

\paragraph{Challenges using LIFT with auxiliary tasks.} Our findings reveal that auxiliary tasks during LIFT offer minimal benefit and can even degrade performance due to overfitting. Additionally, simply fine-tuning the model on long texts does not inherently endow it with robust reasoning capabilities over such texts. These observations underscore the necessity for more effective strategies to harness the in-parameter knowledge of LLMs, enabling them to reason efficiently and accurately over extended contexts.

LIFT is a fascinating concept because humans similarly transform short-term memory into long-term memory, much like LIFT converts in-context knowledge into in-parameter knowledge. While LIFT is far from fully addressing the challenging long-context problem in LLMs, our preliminary results suggest it offers a promising and exciting direction for further research and investment. We encourage the community to explore LIFT with broader training corpora, diverse models, advanced auxiliary task designs, and greater computational resources.

\clearpage

\bibliography{iclr2025_conference}
\bibliographystyle{iclr2025_conference}

\appendix
\clearpage

\section{Experiment Details}
\label{app:details}
\subsection{Hardware settings}

In the experiments of our main results on LooGLE and LongBench (Section \ref{sec:main_results_loogle} and \ref{sec:main_results_longbench}), the LIFT models (i.e., \texttt{LIFT\_only}, \texttt{LIFT+ICL}) were trained and tested on 4 NVIDIA A100 GPUs, and the models w/o LIFT (i.e., \texttt{ICL}) were tested on 1 NVIDIA A100 GPU. In the efficiency test, to ensure fairness in evaluation, we maintained the same hardware setup, utilizing 4 NVIDIA A100 GPUs, in both \texttt{ICL} and \texttt{LIFT}.

In the experiments of our further studies (Section \ref{sec:further_studies}), all the settings with LIFT (i.e., \texttt{LIFT+ICL}, \texttt{LIFT+AT+ICL}, \texttt{SFT+LIFT+ICL}, and \texttt{SFT+LIFT+AT+ICL}) are tested on 4 NVIDIA A100 GPUs and the others are tested on 1 NVIDIA A100 GPUs. Besides, the SFT model is trained on 8 NVIDIA A100 GPUs and the SFT model used in \texttt{SFT+ICL}, \texttt{SFT+LIFT+ICL}, and \texttt{SFT+LIFT+AT+ICL} is the same. It should be noted that 4 NVIDIA A100 GPUs are sufficient to train the SFT model but we adopt 8 GPUs to accelerate training.

In the NIAH tests (Appendix \ref{app:niah}), both models, \texttt{ICL} and \texttt{LIFT+ICL} are tested on 4 NVIDIA A100 GPUs.

\subsection{Hyperparameter settings}

We adopted the same hyperparameter settings of LIFT across all the experiments involving LIFT. The hyperparameters of the segmentation method (Section \ref{subsec:segment}) include $\ell=2048$, which can be fitted in the context window of most short-context models, $s=\frac{3}{8}\ell=768$, and the maximum input length is set to $7900$, which is a bit shorter than the context window of Llama 3 ($8000$) to guarantee the outputs can be fitted in the context window. We adopted full fine-tuning and AdamW in LIFT and the hyperparameters of AdamW are listed in Table \ref{tab:adam_hyperparam}.

\begin{table}[h]
    \centering
    \caption{AdamW Hyperparameter settings}
    \label{tab:adam_hyperparam}
    {\begin{tabular}{lc}
    \toprule
    Hyperparameter & Value  \\
    \midrule
    learning rate & $1.0\times10^{-6}$ \\
    weight decay & $1.0\times10^{-4}$ \\
    max grad norm & $1.0$ \\
    $\beta_{1}$ & $0.9$ \\
    $\beta_{2}$ & $0.98$ \\
    $\epsilon$ & $1.0\times10^{-8}$ \\
    \bottomrule
    \end{tabular}}
    \vspace{-8pt}
\end{table}

Besides, we put all the samples including the context segments and the auxiliary tasks into a single batch through gradient accumulation to stabilize gradients. The actual batch size of a single device is $4$.

\section{Results on Needle-in-a-Haystack (NIAH)}
\label{app:niah}

\begin{figure}[t]
    \centering
    \includegraphics[width=0.8\linewidth]{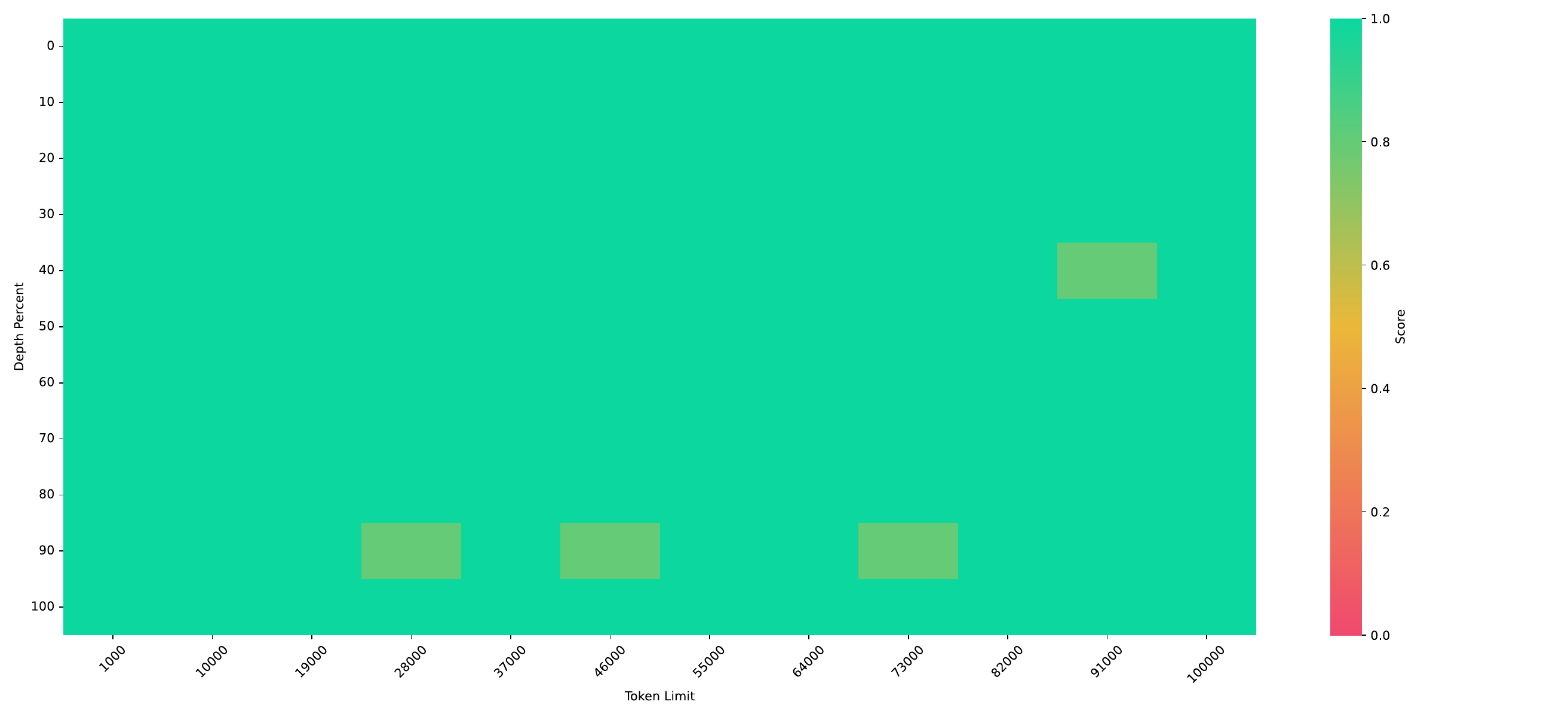}
    \includegraphics[width=0.8\linewidth]{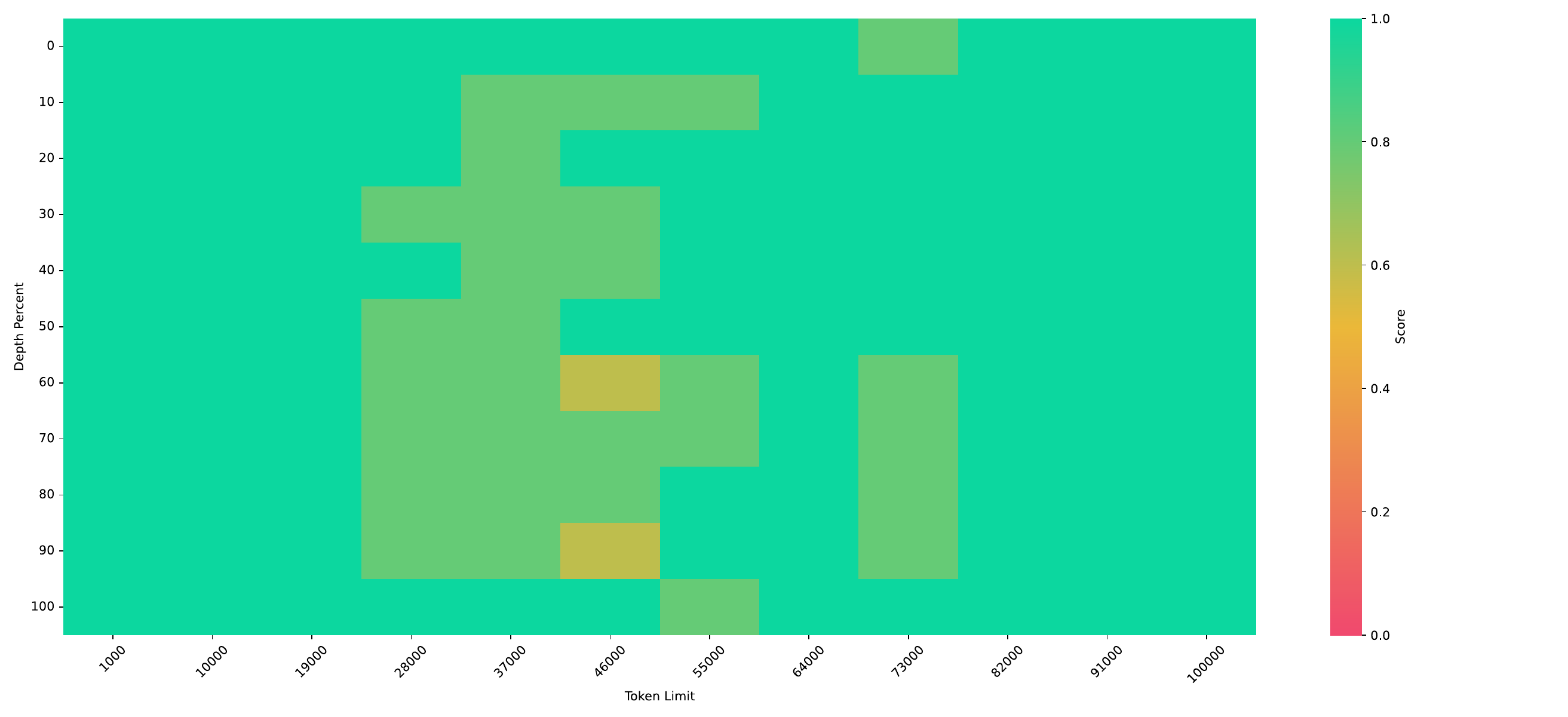}
    \caption{Performance on NIAH: \texttt{ICL} (top) vs. \texttt{LIFT+ICL} (bottom).}
    \label{fig:niah}
\end{figure}

We present the experimental results in the NIAH \citep{niah} task in Figure \ref{fig:niah}, as further analysis of the pros and cons of LIFT and directions for future works. The task requires accurate retrieval from the contexts. We adopt a strong long-context model, LLaMA-3.1-8B-Instruct, as the baseline and apply the LIFT framework to the model.

The maximum context length of our test is 100K, which is within the 128K context window of LLaMA-3.1-8B-Instruct. As expected, the baseline achieves nearly perfect performance. However, LIFT slightly degrades the performance and the degradation seems irregular.

The reason for the degradation may be that LIFT introduces more noise to the model. While most parts of the context are irrelevant to the answer, LIFT asks the model to memorize all the context. The model is likely to be misled by the large amount of irrelevant information.

As summarized in Section \ref{sec:limitation}, precise memorization can be challenging for LIFT. On the one hand, LIFT can't accurately memorize the context while avoiding overfitting. On the other hand, LIFT is likely to be misled when most information is irrelevant to the answer. Future works may improve the LIFT framework from these two aspects.

\section{LIFT can perform much better with extracted evidence}

\begin{table}[b]
    \centering
    \caption{Performance with extracted evidence of Llama3 on LongQA}
    \label{tab:result_evidence}
    {\begin{tabular}{l|ccc}
    \toprule
    Methods & Meteor & Bertscore & GPT4  \\
     \midrule
     \texttt{ICL}  & 9.10 & 83.60 & 30.88 \\
     \texttt{ICL+Evidences} & 11.37 & 83.71 & 53.77  \\
     \texttt{LIFT+ICL} & 9.15 & 83.71 & 33.42 \\
     \texttt{LIFT+ICL+Evidences} & \textbf{12.03} & \textbf{84.06} & \textbf{55.13} \\
    \bottomrule
    \end{tabular}}
    \vspace{-8pt}
\end{table}

\begin{table}[t]
    \centering
    \caption{Performance with extracted evidence of each task in LongQA for Llama3}
    \label{tab:result_evidence_task}
    \resizebox{\linewidth}{!}{\begin{tabular}{l|cccc}
    \toprule
    Methods & Comprehension \& Reasoning
 & Multiple info retrieval
 & Computation & Timeline reorder \\
     \midrule
     \texttt{ICL} & 40.88 & 28.16 & 24.00 & 22.33\\
     \texttt{ICL+Evidences} & 57.14 & \textbf{54.47} & 62.00 & 42.33  \\
     \texttt{LIFT+ICL}  & 44.83 & 28.16 & 22.00 & 26.51  \\
     \texttt{LIFT+ICL+Evidences} &\textbf{57.88} &53.16 & \textbf{64.00} &\textbf{49.30} \\
    \bottomrule
    \end{tabular}}
    \vspace{-8pt}
\end{table}

For a task in LooGLE, the relevant evidences are provided as a sequence of multiple relevant information retrieved from long context for further computation, reorder, reasoning and comprehension to obtain the final answer.

We make further studies on whether extracting relevant evidence can further enhance the long context understanding after LIFT. In Table \ref{tab:result_evidence}, it highlights the effectiveness of integrating evidences and combining it with LIFT in greatly improving the model's performance, which leaves space for further enhancement on the strategy of LIFT. While LIFT alone provides modest improvements, the most substantial gains are observed when evidences are integrated into the ICL process, either with or without LIFT.

Table \ref{tab:result_evidence_task} further expands the performance in Table \ref{tab:result_evidence} on specific tasks in LongQA in LooGLE. \texttt{LIFT+ICL+Evidences} clearly outperforms the other configurations across all metrics, highlighting the importance of extracting relevant knowledge from parameters and executing explicit step-by-step reasoning in more complex tasks like long-dependency QA. The incorporation of evidences helps the model ground its inferences resulting in a more refined and contextually accurate response generation.

\end{document}

%% file: math_commands.tex
\usepackage{amsmath,amsfonts,bm}

\def\eqref#1{equation~\ref{#1}}

\def\1{\bm{1}}

\DeclareMathAlphabet{\mathsfit}{\encodingdefault}{\sfdefault}{m}{sl}
\SetMathAlphabet{\mathsfit}{bold}{\encodingdefault}{\sfdefault}{bx}{n}

